\documentclass[10pt,twocolumn,letterpaper]{article}

\usepackage[pagenumbers]{cvpr} 

\usepackage{graphicx}
\usepackage{amsmath}
\usepackage{amssymb}
\usepackage{booktabs}

\usepackage{array,booktabs,calc}
\usepackage{placeins}
\usepackage{graphicx}
\usepackage{multirow}
\usepackage{wrapfig}
\usepackage{paralist}
\usepackage[dvipsnames]{xcolor}
\usepackage{inconsolata}
\usepackage{gensymb}
\usepackage{caption} 
\usepackage{bbm}  
\usepackage{xcolor}
\usepackage{floatrow}
\usepackage{stfloats}

\newcommand{\PAR}[1]{\vskip4pt \noindent{\bf #1~}}

\newcommand\blfootnote[1]{%
  \begingroup
  \renewcommand\thefootnote{}\footnote{#1}%
  \addtocounter{footnote}{-1}%
  \endgroup
}

\usepackage[pagebackref,breaklinks,colorlinks]{hyperref}

\usepackage[capitalize]{cleveref}
\crefname{section}{Sec.}{Secs.}
\Crefname{section}{Section}{Sections}
\Crefname{table}{Table}{Tables}
\crefname{table}{Tab.}{Tabs.}

\newcommand{\shortname}{SA-HMR\xspace}
\newcommand{\tocite}[1]{\textcolor{red}{[TOCITE]}}

\newcommand{\scene}{\mathcal{S}}
\newcommand{\cmr}{\color{black}\checkmark}

\def\cvprPaperID{2754} 
\def\confName{CVPR}
\def\confYear{2023}

\begin{document}

\title{Learning Human Mesh Recovery in 3D Scenes}

\author{
  Zehong Shen 
  \quad Zhi Cen 
  \quad Sida Peng
  \quad Qing Shuai
  \quad Hujun Bao 
  \quad Xiaowei Zhou$^\dagger$\\[1.5mm]
  State Key Lab of CAD\&CG, Zhejiang University
}
\maketitle

\begin{abstract}
  \textit{
We present a novel method for recovering the absolute pose and shape of a human in a pre-scanned scene given a single image. 
Unlike previous methods that perform scene-aware mesh optimization, we propose to first estimate absolute position and dense scene contacts with a sparse 3D CNN, and later enhance a pretrained human mesh recovery network by cross-attention with the derived 3D scene cues. 
Joint learning on images and scene geometry enables our method to reduce the ambiguity caused by depth and occlusion, resulting in more reasonable global postures and contacts.
Encoding scene-aware cues in the network also allows the proposed method to be optimization-free, and opens up the opportunity for real-time applications.
The experiments show that the proposed network is capable of recovering accurate and physically-plausible meshes by a single forward pass and outperforms state-of-the-art methods in terms of both accuracy and speed.  
Code is available on our project page: \url{https://zju3dv.github.io/sahmr/}. 
}

\end{abstract}

\blfootnote{$^\dagger$Corresponding author.}

\section{Introduction}\label{sec:intro}

\begin{figure}
    \centering
     \includegraphics[width=0.9\linewidth]{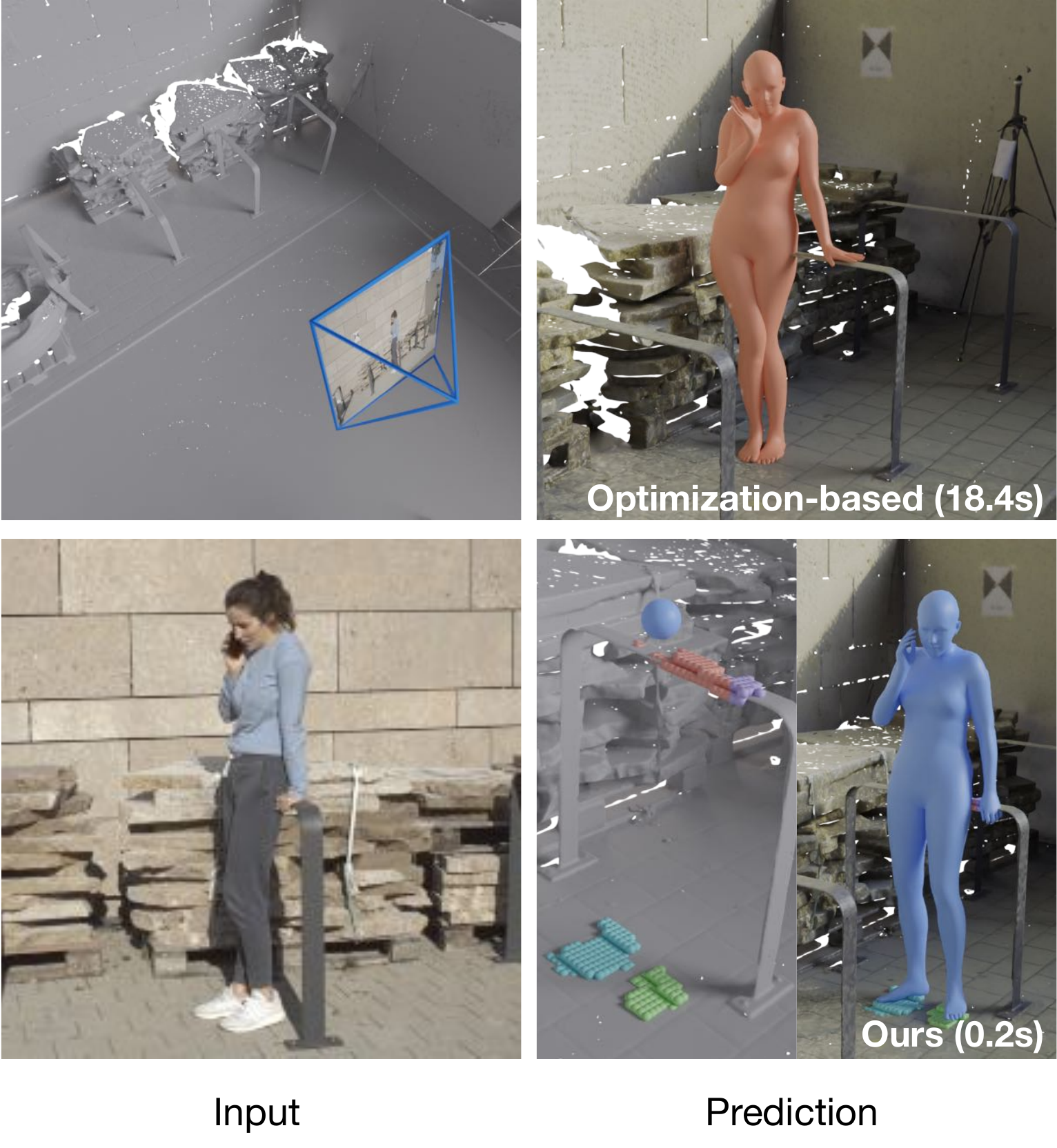}
    \caption{\textbf{Comparison between the optimization-based method and the proposed method.}
    Optimization-based methods typically fit a parametric human model iteratively by minimizing 2D reprojection error and scene conflicts. 
    In contrast, the proposed method utilizes a single forward pass of the network to estimate the global position (blue ball), contact scene points (colored scene points), and a scene-aware human mesh.
    This design leads to improvements in both efficiency and accuracy.
    }
     \label{fig:teaser}
\end{figure}

Monocular human mesh recovery (HMR), i.e., estimating pose and shape parameters of a parametric human model from a single image, 
has gained significant attention in recent years.
To better capture and understand human behaviors, many recent works ~\cite{zhang_lemo_2021,hassan_posa_2021,shimada_hulc_2022,li_mocapdeform_2022,hassan_prox_2019} propose to address the problem of scene-aware HMR which involves human-scene interaction constraints when recovering human meshes, given the 3D geometry of the scene scanned by range sensors~\cite{hassan_prox_2019,huang_rich_2022,zhang2021egobody}, as well as the camera pose of the input image relative to the scene,
which may enable more applications in video surveillance, household robots, and motion analysis in gyms and clinics.

Most existing methods propose using scene-aware optimization to fit the human mesh into a pre-scanned scene.
They optimize a parametric human model iteratively, e.g., SMPL~\cite{loper_smpl_2015}, to minimize scene penetration, chamfer distance of contact regions, and the 3D-2D re-projection error. 
However, optimization tends to be slow at inference time and is sensitive to initialization and hyperparameters, failing to respond in low-latency applications.
As illustrated in Fig.~\ref{fig:teaser}, an optimization-based method PROX~\cite{hassan_prox_2019} takes 18.4s to fit a human model into the scene, while the incorrect position and pose still occur.

Recent works~\cite{kanazawa_hmr_2018,kolotouros_spin_2019,kolotouros_graphcmr_2019,lin_metro_2021,lin_graphormer_2021} propose to recover human mesh with neural networks trained on large-scale datasets~\cite{ionescu_human36m_2014, von2018_3dpw,lin2014microsoft,andriluka20142d}.
Specifically, the networks learn a mapping from an input image to a human mesh in the canonical coordinates. 
Applying these methods in the scene-aware HMR task still requires post-processing optimization, where the global translation and the human poses are refined in accordance with the given scene. 
However, the monocular prediction is conditioned on the input image solely, omitting the joint distribution of human pose and scene geometry, and therefore tends to suffer from depth ambiguity and occlusion. As a result, the optimization-based post-processing could be easily deteriorated by the erroneous initial poses and may even worsen the initial prediction. 

In this work, we propose a Scene-Aware Human Mesh Recovery network (\shortname), 
the first learning-based approach that predicts the absolute position and mesh of a human in the scene by a single forward pass. 
The overall pipeline is illustrated in Fig.~\ref{fig:pipeline}.
Given the input image and scene point cloud, we first use a sparse 3D CNN to estimate dense scene contacts and absolute human position, where the scene contact estimation is treated as a point cloud labeling task, and the human position prediction is presented as a voting vector field refinement task.  
The predicted dense contact points are centered by the human position and passed to a scene network in the human mesh recovery step.
Specifically, we enhance a pretrained monocular HMR network METRO ~\cite{lin_metro_2021} by cross-attention with the proposed scene network in parallel.
In this way, \shortname learns a joint distribution of human pose and scene geometry, resulting in more reasonable postures, contacts, and
global positions, as illustrated in Fig.~\ref{fig:teaser}. Learning scene-aware cues in the network also avoids scene-aware optimization as post-processing and achieves fast inference speed.

We evaluate the proposed method on the RICH~\cite{huang_rich_2022} and PROX~\cite{hassan_prox_2019} datasets of indoor and outdoor scenes.
The experimental results show that \shortname is not only effective in recovering absolute positions and meshes that are in accordance with the given scene, but also significantly faster than the optimization-based baselines.

In summary, we make the following contributions:
\begin{itemize}
\item The first optimization-free framework for scene-aware human mesh recovery from a single image and a pre-scanned scene.
\item The cross-attention design for enhancing a pretrained HMR network with a parallel scene network, enabling joint learning on the human pose and scene geometry.
\item Superior performance compared to optimization-based baselines in terms of both accuracy and speed.
\end{itemize}

\section{Related Work}\label{sec:related}

\PAR{Monocular Human Mesh Recovery.}
Most existing approaches formulate the monocular HMR task as recovering the mesh of statistical human body models, e.g. SMPL and SMPL-X~\cite{loper_smpl_2015,romero_mano_2017,pavlakos_smplx_2019}, where recent works can be divided into optimization-based and learning-based approaches.

The optimization-based approach fits a parametric human model by minimizing the 3D-2D re-projection error of body joints and energy terms of heuristic priors iteratively, which is represented by SMPLify~\cite{bogo_simplify_2016} that fits SMPL~\cite{loper_smpl_2015}.
More recently, SMPLify-X~\cite{pavlakos_smplx_2019} proposes a variational pose prior and fits a more expressive SMPL-X.
Pose-NDF~\cite{tiwari2022posendf} proposes to represent the manifold of plausible human poses with a neural field.
While optimization-based methods are general in their mathematical formulation, they are usually sensitive to hyperparameters and require much time for inference. 

The learning-based methods utilize deep neural networks to predict either the parameters~\cite{kanazawa_hmr_2018,kolotouros_spin_2019,joo_eft_2021,kocabas2021spec} or the mesh vertices~\cite{kolotouros_graphcmr_2019,lin_metro_2021,lin_graphormer_2021} of the SMPL model.
HMR~\cite{kanazawa_hmr_2018} is the pioneering work in predicting SMPL parameters, and SPIN~\cite{kolotouros_spin_2019} improves upon it using an optimization loop.
For predicting SMPL mesh vertices, GraphCMR~\cite{kolotouros_graphcmr_2019} deforms a template human mesh using graph neural network,
while METRO~\cite{lin_metro_2021} uses transformers,
and ~\cite{lin_graphormer_2021} uses graph hierarchy to further improve the performance.
However, for scene-aware HMR, the vertices of the human and scene meshes in contact are close in Euclidean space, making methods that regress parameters unsuitable due to errors that accumulate along the kinematic chains. 
Therefore, the proposed method is built on the works that predict mesh vertices. More details can be found in Sec.~\ref{sec:methods}.

\PAR{Scene-aware Human Mesh Recovery.}
PROX~\cite{hassan_prox_2019} is a seminal work that uses scene constraints to reduce the depth and occlusion ambiguity in monocular HMR.
It achieves this by adding two energy terms of human-scene contact and penetration in the optimization process~\cite{pavlakos_smplx_2019}.
In addition, scene-aware pose generative models~\cite{hassan_posa_2021,zhang_generating_2020} can also be used as prior terms in the scene-aware HMR task.
Other recent works in this area include MoCapDeform~\cite{li_mocapdeform_2022}, which considers deformable scene objects, LEMO~\cite{zhang_lemo_2021}, which uses temporal information, and HULC~\cite{shimada_hulc_2022}, which uses consecutive frames and dense contacts prediction on both the scene and human body.
In contrast to these works, the proposed method is optimization-free and requires only a single forward pass. 

In a broader topic of capturing humans in a scene-aware manner, 
~\cite{shimada_physcap_2020, rempe_humor_2021, luo_embodied_2022} propose using a simulator and dynamic model, where a pre-defined agent is controlled to interact with the scene, and ~\cite{weng2021holistic, zhang_perceiving_2020} consider human-object arrangement by first predicting the human and object and then performing global optimization.

\PAR{Attention in Transformers.}
Attention is a key mechanism in Transformers~\cite{vaswani_transformers_2017}.
It allows a set of query features to fuse the most relevant information from another set of key-value features.
When query and key-value features come from the same source, it is called self-attention, otherwise cross-attention.
In HMR, METRO~\cite{lin_metro_2021} uses self-attention to reduce occlusion ambiguity by establishing non-local feature exchange between visible and invisible parts of a template human mesh.
In feature matching, SuperGlue~\cite{sarlin2020superglue} uses cross-attention to make the corresponding image features more similar.
Predator~\cite{huang2021predator} uses cross-attention in matching two sets of point clouds.

Inspired by feature matching, the proposed method uses cross-attention to potentially make the features of the human and scene that are in contact more similar, resulting in better contact and more reasonable postures.

\section{Methods}\label{sec:methods}

\begin{figure*}
    \centering
     \includegraphics[width=.95\linewidth]{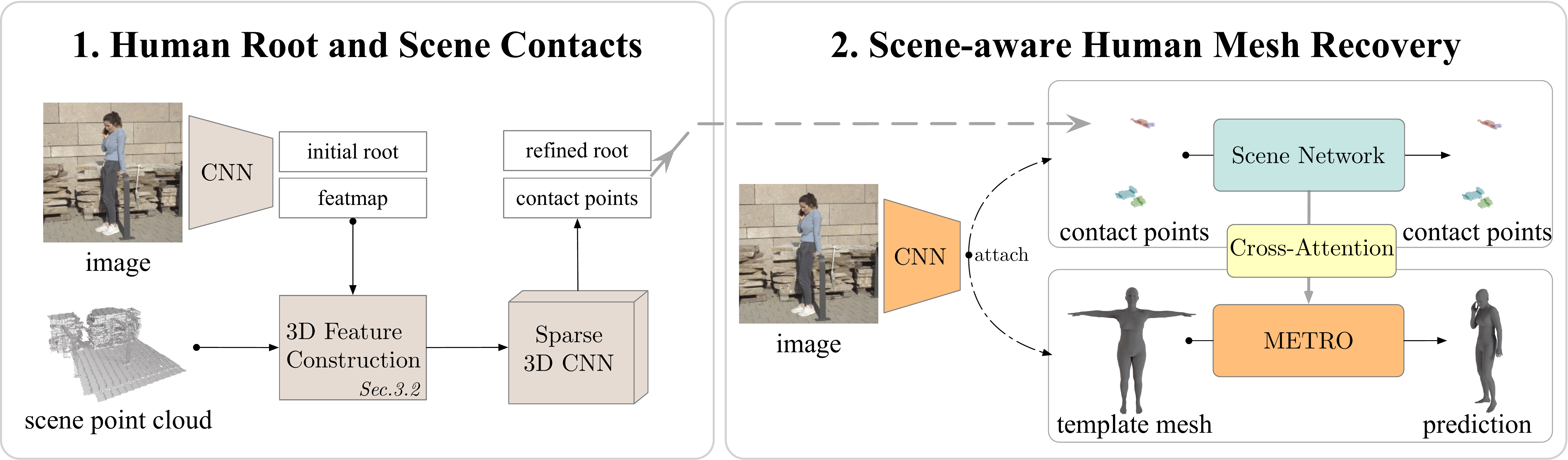}
     \caption{
        \textbf{Overview of the proposed \shortname.} 
     \textbf{1.} The human root and scene contact estimation module (Sec.~\ref{sec:root_contact}) that first predicts the initial root and then refines the root with 3D scene cues using a sparse 3D CNN. 
     The module also predicts contact labels~\cite{hassan_prox_2019} for each scene point.
     Please refer to Sec.~\ref{sec:root_contact} for a detailed definition of the 3D feature construction module. 
     \textbf{2.} The scene-aware human mesh recovery module (Sec.~\ref{sec:sahmr}) that enhances the pretrained METRO~\cite{lin_metro_2021} network with a parallel scene network.
     The scene network takes the predicted contact scene points as input, and uses cross-attention to pass messages to the intermediate features of the METRO network.
     }
     \label{fig:pipeline}
\end{figure*}

Given a calibrated image and a pre-scanned scene point cloud (Sec.~\ref{sec:preliminary}), \shortname first estimates the absolute human root position and scene contacts (Sec.~\ref{sec:root_contact}), and then recovers the human mesh with the contact points by enhancing a pretrained METRO network (Sec.~\ref{sec:sahmr}). 
An overview of the proposed method is presented in Fig.~\ref{fig:pipeline}.

\subsection{Preliminaries}\label{sec:preliminary}

\PAR{Human Representation.} 
We use SMPL~\cite{loper_smpl_2015} as the human representation. 
The SMPL is a parametric model that uses the body joint rotations, root translation, and body shape coefficients to compute the body mesh. 
Following ~\cite{kolotouros_graphcmr_2019, lin_metro_2021}, we directly predict the SMPL mesh vertices $V=\mathbb{R}^{6890\times 3}$, and use H36M~\cite{ionescu_human36m_2014} joint regression matrix $M\in\mathbb{R}^{14\times 6890}$ to compute 3D joints $J\in\mathbb{R}^{14\times 3}$ from the vertices for quantitative evaluation, $J=MV$.

\PAR{Scene and Image Representation.}
We assume that the scene is pre-scanned with range sensors, as in RICH~\cite{huang_rich_2022} and PROX~\cite{hassan_prox_2019}, 
and the image is calibrated and localized in the scene, i.e. with known intrinsic and extrinsic parameters $\{(f,c_x,c_y), (R_c,t_c)\}$. 
Following METRO~\cite{lin_metro_2021}, we detect a squared bounding box around the target human and resize the cropped region as the input image $I\in\mathbb{R}^{224\times224\times3}$. 
Based on the camera parameters and the bounding box, we select scene points that fall within the visual frustum as the input scene point cloud $S\in\mathbb{R}^{N_S\times3}$.

\PAR{Human-Scene Contact.}
Following PROX~\cite{hassan_prox_2019}, we use 7 regions of the SMPL mesh that are most likely to be contacted. The details are provided in the supplementary material.
Using these 7 categories and one for not being in contact, we perform a segmentation task on the scene point cloud.

\subsection{Human Root and Scene Contacts}\label{sec:root_contact}
Given an image $I$ bounding the human, we propose using a 2D convolutional neural network (CNN) to extract image features $F$ and predict the initial human root $r$.
Based on the scene points $S$, we unproject the image features $F$ to 3D, resulting in $\hat F$.
Additionally, we calculate point-wise offset vectors $O$ that point from a voxelized scene point cloud to the initial root.
By taking $\hat F$ and $O$ as input, a sparse 3D CNN predicts the segmentation of scene contacts and the refined offsets, which are then converted to the refined human root $r^*$.
An overview of this process is presented in the left column of Fig.~\ref{fig:pipeline}.

\PAR{Initial Root.}
We predict the initial root $r\text{=}(X,Y,Z)$ in a 2.5D manner following SMAP~\cite{zhen_smap_2020}.
Specifically, we use a CNN to predict the 2D heatmap and a normalized depth map of the root.
Then, the 2D position $(x, y)$ is obtained by applying \texttt{argmax} to the heatmap, and the corresponding normalized depth value $\tilde{Z}$ is retrieved from the depth map.
Finally, using the intrinsic parameters ${f,c_x, c_y}$ and image size $w$, the 3D root position is computed:
\begin{equation}
    Z=\tilde{Z}\frac{f}{w}
\end{equation}
\begin{equation}
    X=\frac{x-c_x}{f}\cdot Z,\quad 
    Y=\frac{y-c_y}{f}\cdot Z
\end{equation}

From our observation, estimating $(x, y)$ achieves good results with a mean squared error of fewer than two pixels across datasets. 
However, estimating $\tilde{Z}$ is relatively challenging, possibly due to the variations in human shapes.

\begin{figure}
    \centering
     \includegraphics[width=0.8\linewidth]{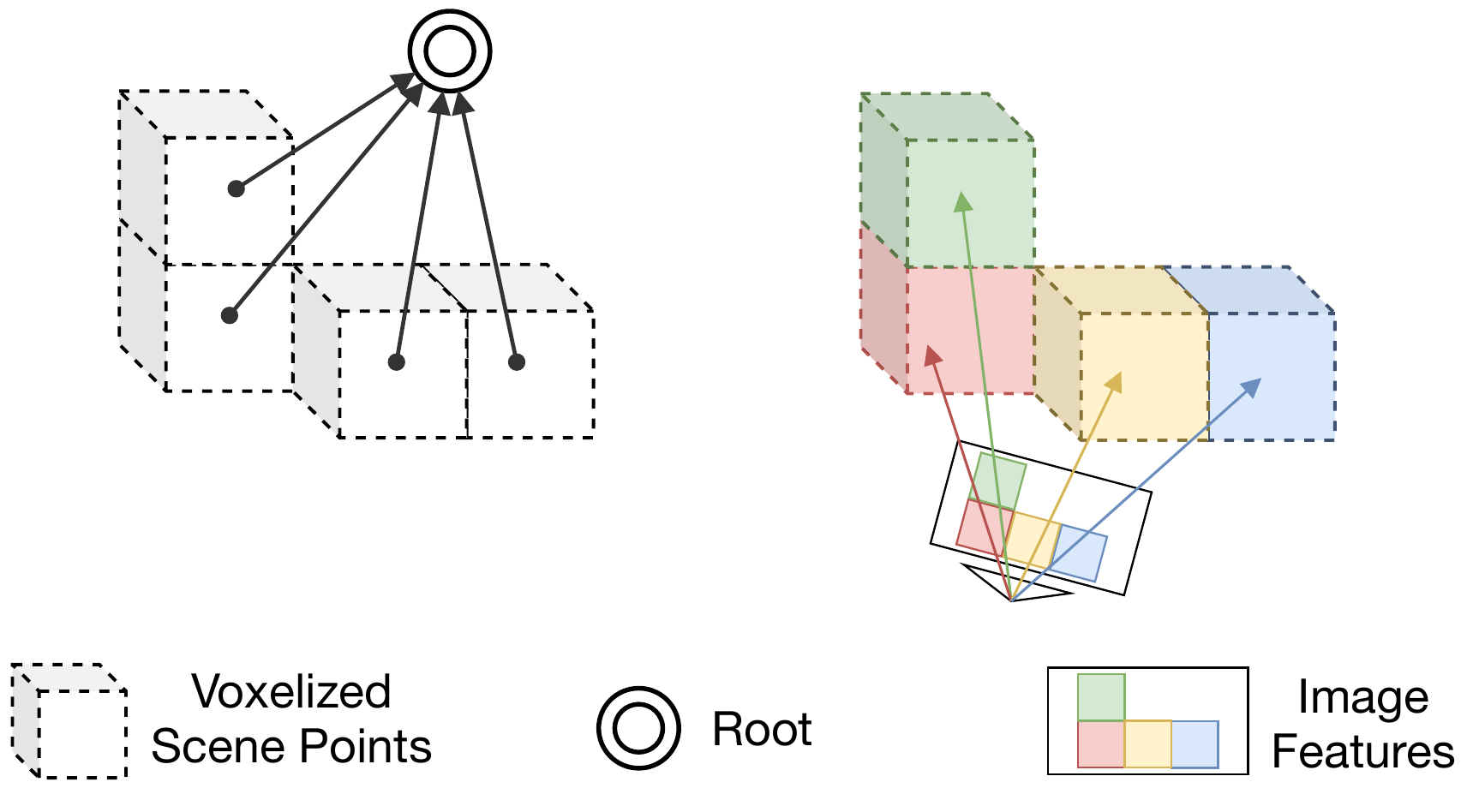}
     \caption{
        \textbf{3D feature construction.} We voxelize a scene point cloud to a sparse volume. 
        The initial feature of each voxel consists of two parts,
        which are the offset vector pointing from the voxel center to the human root and the unprojected image features.  
     }
     \label{fig:3d_feat_construct}
\end{figure}

\PAR{3D Feature Construction.}
Based on the initial root $r$ and image features $F$, we construct the 3D features on the voxelized scene point cloud, which is illustrated in Fig.~\ref{fig:3d_feat_construct}.

First, we select regions of interest around the initial root $r$ in the point cloud.
Specifically, we treat $r$ as an anchor and keep points within a radius $\gamma_1$.
Since $Z$ has more uncertainty than $(x, y)$, we sample two additional anchors along the z-axis, whose distance to $r$ is $\gamma_2$.
Next, we construct a sparse volume $\bar{S}$ by voxelizing these points with voxel size $s_{vox}$, where the center of each voxel is denoted as $\bar {s}_i$.

For each voxel $i$, the feature consists of the offset vector ${o}_i$ and the unprojected image feature $\hat{f}_i$. Specifically, $o_i$ is a vector pointing from the voxel center to the human root:
\begin{equation}
    {o}_{i} = r - \bar{s}_i
\end{equation}
$\hat f$ is computed by projecting voxel center $\bar{s}_i$ onto the image using camera parameters and bilinearly sampling the image feature map $F$.

\PAR{Estimating Refined Root and Scene Contacts.}
We use sparse 3D CNN~\cite{tang2020spvcnn} to process the constructed 3D features and learn to improve the root estimation and predict the scene contacts.
Specifically, the output of each voxel includes an updated offset vector ${o}^*_i$, confidence $c_i$, and segmentation indicating the contact category.
We compute the refined root $r^*$:
\begin{equation}
    r^* =  \sum_{i}c_i\cdot(o^{*}_{i}+\bar{s}_i).
\end{equation}
There are 8 categories of contact points, including 7 most probable regions on the body that would be contacted~\cite{hassan_prox_2019} and 1 category of not being in contact. 
We take the category of the highest score as the prediction for the voxel and set the dense point cloud belonging to the voxel with that category.
The contact points $\hat S_{seg3d}\in \mathbb{R}^{\hat{N}_{S}\times 3}$ serve as the input for the mesh recovery module.

\subsection{Scene-aware Human Mesh Recovery}\label{sec:sahmr}

\begin{figure}
    \centering
     \includegraphics[width=0.95\linewidth]{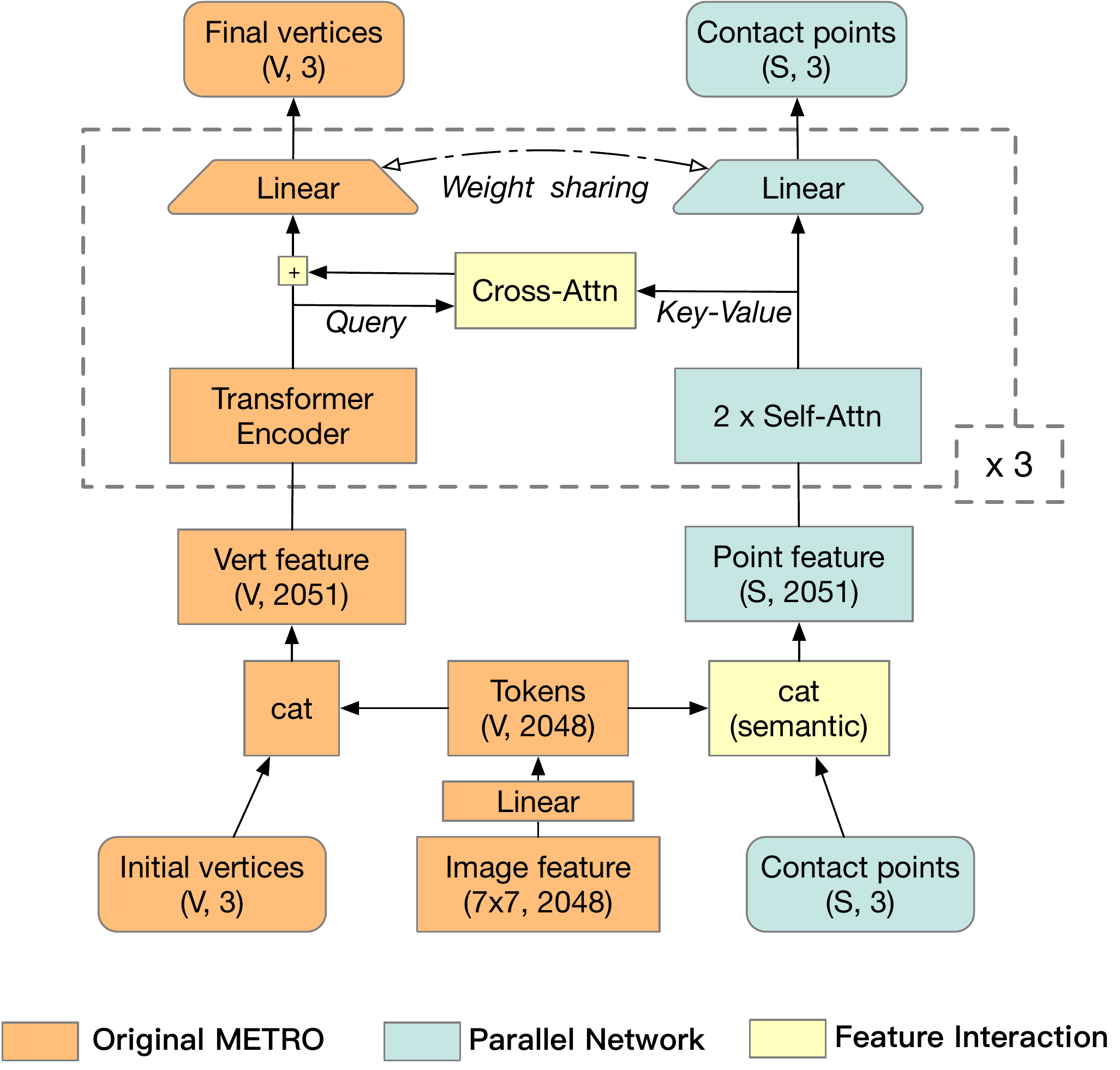}
     \caption{\textbf{Enhancing METRO with a parallel network.} 
     The orange parts are the original METRO~\cite{lin_metro_2021} network, where the residual connection and positional encoding are omitted for simplicity. 
     The blue parts are the proposed parallel network which takes predicted contact scene points as input.
     The yellow parts indicate feature interaction between METRO and the parallel network.}
     \label{fig:enhance_metro}
\end{figure}

Since the training data of scene-aware human mesh recovery is limited, we build our model upon a network named METRO~\cite{lin_metro_2021} that is pre-trained on large-scale data of monocular human mesh recovery.
METRO processes feature based on the self-attention mechanism, and our approach enhances METRO by adding a parallel scene network, which provides a cross-attention-based mechanism that enables METRO to notice important scene details and achieve scene-aware human mesh recovery.

\PAR{METRO} consists of a CNN backbone and multiple Transformer encoders. It first extracts global CNN features, then combines the feature to the vertices of a zero posed SMPL mesh, and finally predicts a posed mesh with shape through the transformers. The part of the transformer is illustrated in the orange part of Fig.~\ref{fig:enhance_metro}.

\PAR{Enhancing METRO with Cross-Attention.}
We improve METRO with a scene network, which makes the predicted human vertices to be close to the corresponding contact scene points $\hat S_{seg3d}$ (Sec.~\ref{sec:root_contact}).

As illustrated in Fig.~\ref{fig:enhance_metro}, we add a parallel network, which has a similar architecture as the transformer of METRO, to extract features of scene contact points and output the point positions like an autoencoder.
Specifically, we first use METRO's CNN backbone to extract the image feature and map it to a set of vertex tokens by a fully-connected layer.
In METRO, these tokens are directly concatenated with the positions of initial human vertices.
However, the number of scene contact points is not the same as the tokens', and there is no one-to-one correspondence between them.
To resolve this issue, we perform the average pooling to vertex tokens based on the contact categories defined on the SMPL mesh vertices~\cite{hassan_prox_2019},
resulting in 7 tokens.
Then, we append each scene contact point with the corresponding aggregated token based on the predicted category in Sec.~\ref{sec:root_contact}.
Intuitively, this helps the cross-attention to focus on the semantically corresponding parts.
To be invariant to the global translation, the scene contact points are zero-centered by the predicted root $r^*$.

\begin{figure}
    \centering
    \includegraphics[width=0.9\linewidth]{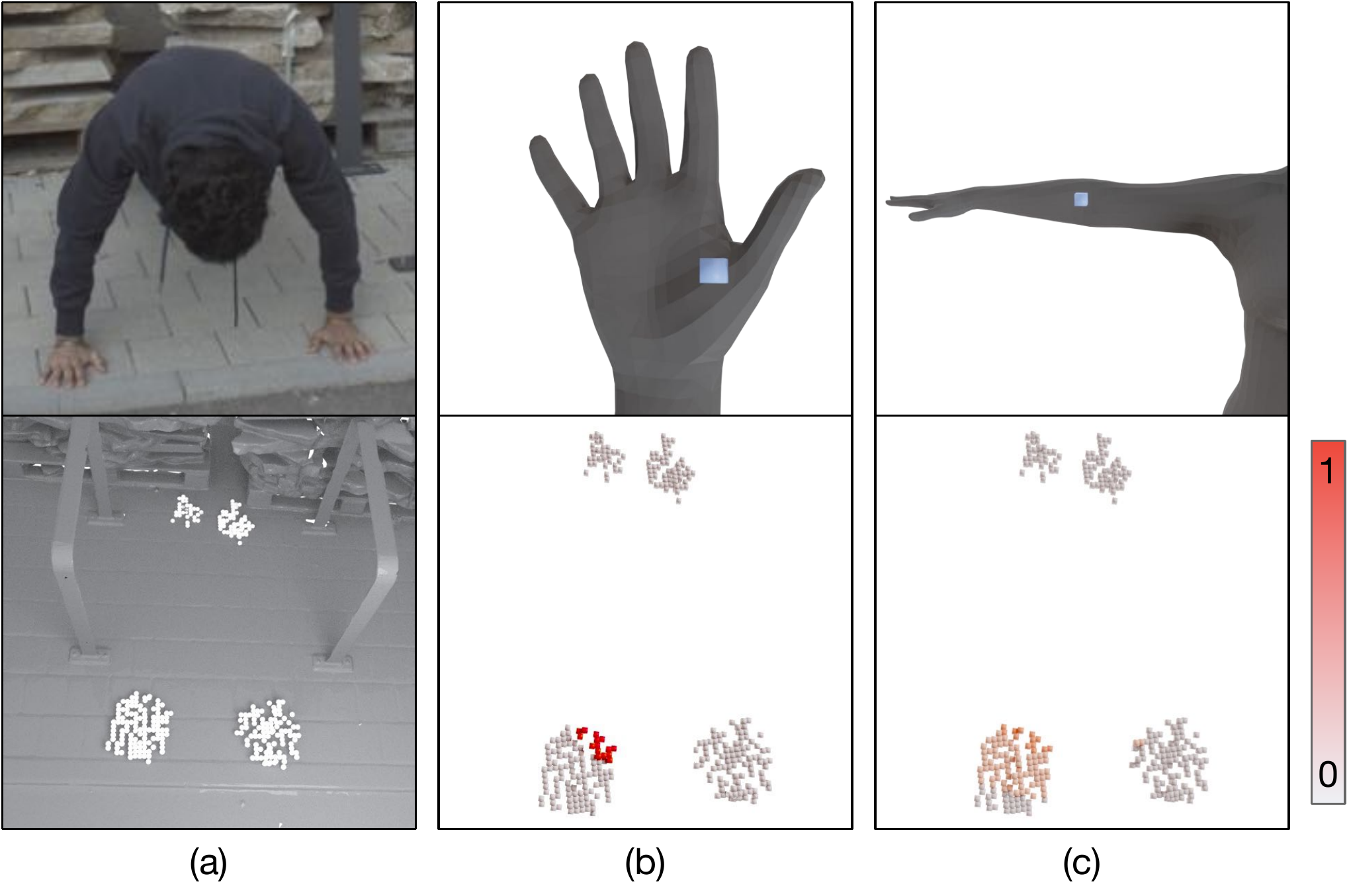}
     \caption{\textbf{Visualization of the cross-attention from a body vertex (blue point) to the predicted dense scene contacts (white points).}
     (a) The image and the predicted dense scene contacts.
     (b) The hand vertex is in contact with the scene according to the image, and its feature is similar to the nearby scene point features, enabling the final vertex prediction to be close to the corresponding scene surface.
     (c) The arm vertex is not in contact with the scene, where feature similarities tend to be evenly distributed. 
     The feature similarities are normalized to the same range of 0$\to$1.  
     }
    \label{fig:cross_attn}
\end{figure}

Motivated by recent feature matching methods~\cite{sarlin2020superglue,huang2021predator}, we propose to use cross-attention to pass features from scene contact points to human vertices.
The cross-attention and self-attention share the same underlying mechanism, both of which first compute the similarity of query and key, and then use the weighted sum to fuse features. 
When the query and key are from the same source features, it is self-attention, and otherwise is cross-attention.
In practice, we use linear attention operator~\cite{katharopoulos2020linear} to improve efficiency.
A visualization of cross attention over human vertices and scene points is in Fig.~\ref{fig:cross_attn}. 
The detailed network architecture is provided in the supplementary material.

Note that, we use a weight-sharing regressor layer with METRO, which regresses from point-wise features to point position $(x,y,z)$. Therefore, to get a similar $(x,y,z)$, the input features of this layer should also be similar.
This strategy implicitly aligns the features of human vertices and scene points when their final predictions are near in 3D space, thus facilitating the cross-attention to find correspondences between human vertices and scene points.

\subsection{Training Loss}\label{sec:training_loss}
\PAR{Root and Contact.}
The loss function $L_{\text{RC}}$ for the root and contact estimation is defined as:
\begin{equation}
    L_{\text{RC}} = L_{\text{R2D}} + w_{\text{RZ}}\cdot L_{\text{RZ}} + L_{\text{ROV}} + L_{\text{R3D}}  + L_{\text{C}}
\end{equation}
where $L_{\text{R2D}}$ is the MSE loss on the root heatmap;
$L_{\text{RZ}}$, $L_{\text{ROV}}$, and $L_{\text{R3D}}$ are the L1 losses on the relative depth, offset vectors, and the 3D root, respectively;
$L_{\text{C}}$ is the cross-entropy loss for contact categories of voxel points, where we additionally train an auxiliary task of 2D contact segmentation similar to the voxel points.

\PAR{Human Mesh Recovery.}
The loss function $L_{\text{HMR}}$ for the mesh recovery is defined as:
\begin{equation}
    L_{\text{HMR}} = L_{\text{V}} + L_{\text{J}} + L_{\text{CP}} + L_{\text{GV}} 
\end{equation}
where $L_{\text{V}}$, $L_{\text{J}}$, $L_{\text{CP}}$, and $L_{\text{GV}}$ are the L1 losses on the translation-aligned human vertices, human joints, reconstructed contact points, and global human vertices, respectively. More details are in the supplementary material.

\subsection{Implementation Details}\label{sec:impl_details}
We train two modules separately.
For the root and contact module,
the CNN is HRNet-stride-4~\cite{sun_deep_2019} with METRO initialization,
the sparse 3D CNN is SPVCNN~\cite{tang2020spvcnn, tang2022torchsparse} with random initialization.
We use linear layers to align intermediate feature dimensions. 
$\gamma_1$ is $1.25m$, $\gamma_2$ is $0.5m$, $s_{vox}$ is $5^3cm^3$, and $w_{\text{RZ}}$ is 10.
The contact threshold is $7cm$.
We flip images for augmentation.
The module is trained with an initial learning rate of 3.75e-5 and a batch size of 24. 
It converges after 30 epochs of training on one V100 GPU.
For the mesh recovery module, the METRO network is initialized as pretrained and the scene network is randomly initialized. 
The initial learning rate is 7.5e-6 and the batch size is 24. 
It converges after 30 epochs of training.

\section{Experiments}\label{sec:exp}

\begin{table*}
    \centering
    \scriptsize{\resizebox{0.95\textwidth}{!}{%
    \setlength\tabcolsep{4.0pt}
    \begin{tabular}{lccccccccc}
        \toprule
        Method & Learning-based & Optimization & Scene-aware & G-MPJPE$\downarrow$ & G-MPVE$\downarrow$ & PenE$\downarrow$ & ConFE$\downarrow$ & MPJPE$\downarrow$ & MPVE$\downarrow$ \\
        \midrule
        Dataset GT~\cite{huang_rich_2022} &&\cmr&\cmr       & /&/&~9.8&10.8&/&/\\
        \midrule
        SMPLify-X~\cite{pavlakos_smplx_2019} & &\cmr&       &482.0&483.7&35.7&43.4&166.9&177.6 \\
        PROX~\cite{hassan_prox_2019} & &\cmr&\cmr           &\bf{390.1}&\bf{397.2}&15.5&24.1&164.1&175.8 \\
        POSA~\cite{hassan_posa_2021} & &\cmr&\cmr           &427.8&434.0&21.1&27.0&177.2&188.4 \\
        PLACE~\cite{zhang_place_2020} & &\cmr&\cmr          &395.9&403.0&16.1&24.8&163.8&175.4 \\
        METRO~\cite{lin_metro_2021}$^\dag$~+ SA-Opt~\cite{huang_rich_2022, zhang_perceiving_2020}
                            ~&\cmr&\cmr&\cmr                &563.1&561.3&\bf{~7.4}&\bf{14.8}&\bf{102.7}&\bf{112.8} \\
        \midrule
        METRO~\cite{lin_metro_2021} &\cmr& &                &678.6&679.4&52.2&56.9&129.6&134.5 \\
        METRO~\cite{lin_metro_2021}$^\dag$ &\cmr& &         &511.7&509.7&33.6&37.6&~98.8&107.9 \\
        \textbf{\shortname} &\cmr& &\cmr                    &\bf{264.6}&\bf{272.7}&\bf{14.9}&\bf{19.0}& ~\bf{93.9}&\bf{103.0} \\
        \bottomrule
    \end{tabular}
}}
    \caption{
    \textbf{Evaluation on the RICH~\cite{huang_rich_2022} dataset.}
    METRO$^\dag$ indicates that the model is finetuned on the dataset.
    SA-Opt indicates scene-aware optimization, with contact estimation from BSTRO~\cite{huang_rich_2022} and loss formulation from PROX~\cite{hassan_prox_2019} and PHOSA~\cite{zhang_perceiving_2020}.
    The proposed \shortname achieves the overall best results and is significantly faster than the methods that require optimization.
    }
    \label{tab:rich_test}
\end{table*}

\subsection{Datasets}
We train and evaluate the proposed method on RICH~\cite{huang_rich_2022} and PROX~\cite{hassan_prox_2019} datasets separately.

\PAR{RICH~\cite{huang_rich_2022}} captures multi-view video sequences in 6 outdoor and 2 indoor environments.
It provides images, reconstructed bodies, scene scans, and human-scene contact labels annotated on SMPL vertices.
We skip frames including multiple subjects, remove the first 45 frames of each video to avoid static starting pose, and skip frames where the subjects' 2D bounding boxes are not inside the images.
Then we downsample the train / val / test splits to 2 / 1 / 1 fps, resulting in 15360 / 3823 / 3316 frames. 

\PAR{PROX~\cite{hassan_prox_2019}} captures monocular RGBD videos in 12 indoor environments.
We use RGB images and scene scans.
It is a challenging dataset where severe occlusions exist in most frames.
We use the qualitative set for training and the quantitative set for testing.
In order to get better training annotations, we additionally combine HuMoR~\cite{rempe_humor_2021} which utilizes motion prior and optimizes a sequence of frames.
Then, we manually remove the failed frames that are not consistent with the images and scenes.
Finally, the training split contains 4852 frames.

\subsection{Metrics}\label{sec:metrics}
We evaluate quantitatively in terms of human mesh recovery and human-scene contact.

\PAR{Human Mesh Recovery.}
We report the Global Mean-Per-Joint-Position-Error (\textbf{G-MPJPE}) and Global Mean-Per-Vertex-Error (\textbf{G-MPVE}) in scene coordinates,
which calculates the average L2 distances between predicted and ground truth joints/vertices.
Additionally, we report the translation-aligned metrics \textbf{MPJPE} and \textbf{MPVE}.

\PAR{Human-scene Contact.} 
We report the Penetration Error (\textbf{PenE}) and Contact Failure Error (\textbf{ConFE}).
PenE measures the total distance that SMPL vertices penetrate the scene mesh:
\begin{equation}
    \text{PenE} = \sum_{i=1}^{V} \mathbbm{1}_{x<0} [sdf(v_i, \scene)] \cdot |sdf(v_i, \scene)|,
\end{equation}
where $V$ is the number of SMPL vertices, 
$sdf\left(v_i, \scene\right)$ is the signed distance of vertex $v$ to scene $\scene$,
and $\mathbbm{1}_{x<0} [\cdot]$ is an indicator function that returns 1 when the condition is met, and 0 otherwise.
ConFE measures contact quality when the ground-truth contact label is available:
\begin{equation}
    \begin{split}
        \text{ConFE} = &\sum_{i=1}^{V}(C_{gt}(v_i) \cdot |sdf(v_i, \scene)| \\
                     + &(1-C_{gt}(v_i)) \cdot \mathbbm{1}_{x<0} [sdf(v_i, \scene)] \cdot | sdf(v_i, \scene)|), \\
    \end{split}
\end{equation}
where $C_{gt}(v)$ equals 1 if $v$ is labeled as in contact, and 0 otherwise.
In order to obtain a good result of ConFE, the body vertices in contact should be near the scene surface, while vertices not in contact should avoid penetration.

\subsection{Main Results}\label{sec:results}

\begin{table}
    \centering
    \scriptsize{

\resizebox{1.0\textwidth}{!}{%
    \setlength\tabcolsep{4.0pt}
    \begin{tabular}{lccccc}
        \toprule
        Method  & G-MPJPE$\downarrow$ & G-MPVE$\downarrow$ & PenE$\downarrow$ & MPJPE$\downarrow$ & MPVE$\downarrow$ \\
        \midrule
        Dataset GT~\cite{hassan_prox_2019}      &/&/&~9.6&/&/\\
        \midrule
        SMPLify-X~\cite{pavlakos_smplx_2019}    &216.0&222.6&49.3&\bf{100.7}&\bf{112.8} \\
        PROX~\cite{hassan_prox_2019}            &172.0&178.5&\bf{10.7}&101.1&114.0 \\
        POSA~\cite{hassan_posa_2021}            &172.3&180.9&16.6&108.5&119.4 \\
        PLACE~\cite{zhang_place_2020}           &\bf{168.1}&\bf{176.7}&12.3&100.8&113.7 \\
        \midrule
        METRO~\cite{lin_metro_2021}             &283.2&277.7&62.4&137.0&147.2 \\
        METRO~\cite{lin_metro_2021}$^\dag$      &265.6&262.7&67.5&117.1&128.5 \\
        \textbf{\shortname}                     &\bf{150.4}&\bf{160.0}&\bf{26.9}&\bf{111.1}&\bf{122.5} \\
        \bottomrule
    \end{tabular}
}}
    \caption{\textbf{Evaluation on the PROX~\cite{huang_rich_2022} dataset.}
    The proposed method achieves the best performance in global metrics. 
    }\label{tab:prox}
\end{table}

\PAR{Baselines.} 
Optimization: 
SMPLify-X~\cite{pavlakos_smplx_2019} uses RGB only, 
PROX~\cite{hassan_prox_2019} extends it with losses of human-scene contact and penetration, 
POSA~\cite{hassan_posa_2021} and PLACE~\cite{zhang_place_2020} extend PROX with scene-aware pose priors,
and METRO~\cite{lin_metro_2021}$^\dag$+SA-Opt stands for post-processing a finetuned METRO with scene-aware optimization, which will be explained later.
Learning-based:
METRO predicts canonical human mesh vertices and a weak-perspective camera.
We solve the transformation from human to camera coordinate by minimizing joint re-projection error with a PnP solver~\cite{opencv_library}.
The METRO$^\dag$ is finetuned with the same training protocol as \shortname.
Since METRO does not consider scenes, we additionally optimize the global pose and scale by minimizing scene-aware losses, including re-projection error, human-scene penetration, contact distance~\cite{huang_rich_2022}, and ordinal depth error~\cite{zhang_perceiving_2020}, following the key ideas of PROX and PHOSA~\cite{zhang_perceiving_2020}.
 
\begin{figure*}
    \centering
     \includegraphics[width=0.96\linewidth]{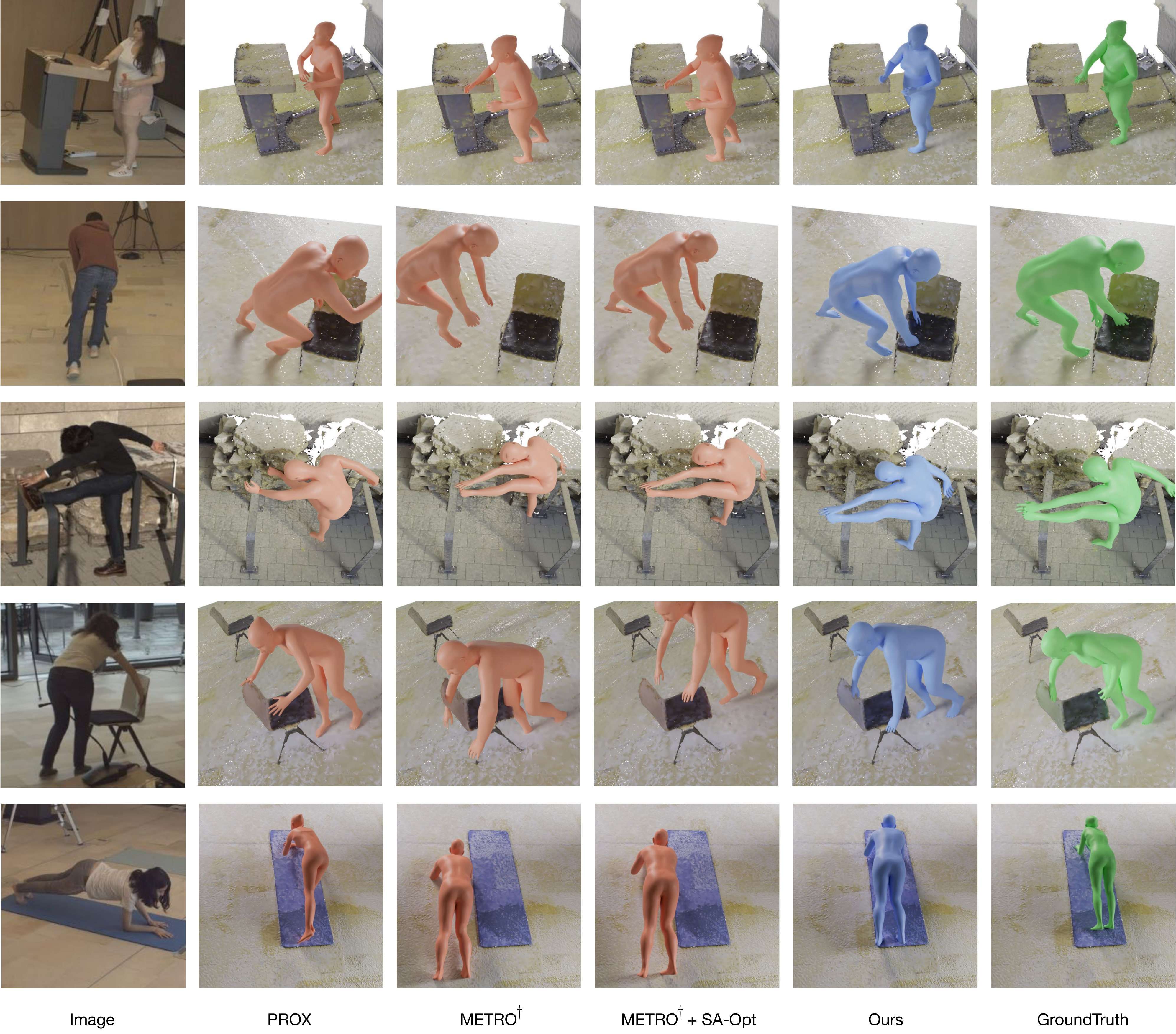}
     \caption{\textbf{Qualitative results on the RICH~\cite{huang_rich_2022} dataset.}
    We compare the proposed method to PROX~\cite{hassan_prox_2019}, finetuned METRO~\cite{lin_metro_2021}, finetuned METRO with scene-aware optimization,
    and ground truth.
    The leftmost column shows the input images.
    The proposed method recovers the global positions and human-scene contact more accurately because of the 3D learning on human root refinement and dense scene contact labeling tasks.
    }
     \label{fig:qual}
\end{figure*}

\PAR{Results.}
For the RICH dataset, Tab.~\ref{tab:rich_test} shows that \shortname notably outperforms other baselines in terms of the G-MPJPE and G-MPVE by a significant margin, demonstrating the effectiveness of the proposed pipeline.
The joint learning on both image and scene geometry also improves the metrics of local pose and human-scene contact.
We use open-sourced code for SMPLify-X and PROX, and implement POSA and PLACE upon PROX.
Optimization methods approximately cost 18s for a single fitting, which is much slower compared to 0.2s of \shortname.
We provide qualitative results comparing to baselines in Fig.~\ref{fig:qual}. 

For the PROX dataset, \shortname outperforms all baselines in terms of global accuracy as illustrated in Tab.~\ref{tab:prox}. 
Since the pseudo ground truth is still not of low quality for the training set of PROX, as well as a domain gap exists in the test set where the subject wears a MoCap suit, our method falls a little behind in local accuracy and scene penetration.
And we do not report ConFE, since the ground-truth contact label is not available.
Nevertheless, the clear improvement compared to the most relevant model METRO$^\dag$ has demonstrated the effectiveness of the proposed method. 

We also observe that while considering the scene geometry is critical for estimating the global position and improving physical plausibility, it may not fully resolve the ambiguity of the local pose, where multiple physically plausible solutions may still exist.
For example, the RGB-only method SMPLify-X and the scene-aware method PROX perform similarly in MPJPE and MPVE.

\subsection{Ablation Study}\label{sec:abl_study}

\begin{table}
    \centering
    \scriptsize{\resizebox{0.75\textwidth}{!}{%
    \setlength\tabcolsep{4.0pt}
    \begin{tabular}{lccc}
        \toprule
        Dataset & Initial RtErr & Refined RtErr & Final RtErr \\
        \midrule
        RICH	 & 510.8 & 284.7 & \textbf{246.5} \\
        PROX	 & 364.2 & 132.3 & \textbf{111.8} \\
        \bottomrule
    \end{tabular}
}

}
    \caption{
        \textbf{Ablation study of root estimation.}
        The human root position errors (RtErr) in mm are reported.
    }\label{tab:abl_root}
\end{table}

\begin{figure*}
    \centering
    \includegraphics[width=0.96\linewidth]{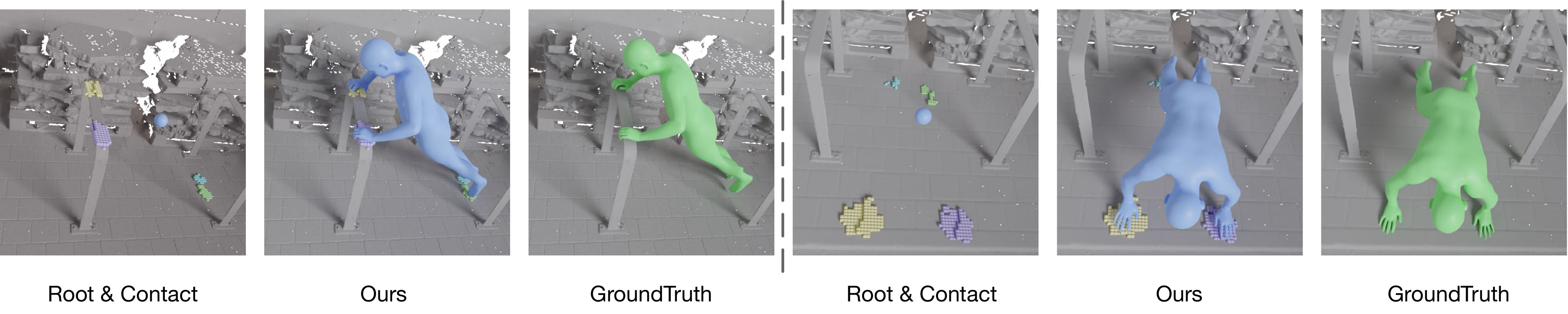}
    \caption{\textbf{Qualitative visualization of the estimated root locations and dense scene contacts.} 
    In both examples, the estimated contact points provide accurate position and scene structure for the following step of mesh recovery.
    The reconstructed human mesh is in good contact with the corresponding scene regions. 
    }
    \label{fig:qual_contact}
\end{figure*}

\PAR{Root and Contact Module.}
Tab.~\ref{tab:abl_root} shows that the predicted human root position is improved progressively by the refinement and scene-aware HMR modules,
where the initial prediction~\cite{zhen_smap_2020} is improved 44\%/52\% in RICH, and 64\%/69\% in PROX. 
The offset representation helps to improve erroneous initial root prediction that is not consistent with the scene surface.
For scene contact estimation, the precision/recall is 0.57/0.53 on RICH, and 0.45/0.24 on PROX. 
We observe that the contacts are difficult to predict, which aligns with the conclusion of a recent work HULC~\cite{shimada_hulc_2022}.
More visualizations are presented in Fig.~\ref{fig:qual_contact}. 

\begin{table}
    \centering
    \scriptsize{\resizebox{1.\textwidth}{!}{%
    \begin{tabular}{lccccc}
        \toprule
        Method & G-MPJPE & G-MPVE & MPJPE & MPVE & CErr$\downarrow$\\
        \midrule
        w/o parallel	& 304.8 & 312.9 & 98.5 & 108.9 & 10.2\\
        Ours	        & \textbf{264.6} & \textbf{272.7} & \textbf{93.9} & \textbf{103.0} & ~\textbf{8.9}\\
        \bottomrule
    \end{tabular}
}

}
    \caption{\textbf{Ablation study of the parallel network} on the RICH dataset.
    The compared variant fuses features of contact points at the early stage. 
    }\label{tab:abl_parallel}
\end{table}

\begin{table}
    \centering
    \scriptsize{
\resizebox{0.65\textwidth}{!}{%
    \setlength\tabcolsep{4.0pt}
    \begin{tabular}{ccccc}
        \toprule
        Root & Contact & MPJPE & MPVE & CErr$\downarrow$ \\
        \midrule
        / &   /& 98.8  & 107.9 & 10.5   \\
      Est.&Est.& 93.9  & 103.0 & ~8.9   \\
        GT&Est.& 89.2  & 98.1  & ~7.9   \\
      Est.& GT & 90.4  & 99.2  & ~8.3     \\
        GT& GT & \bf{76.7} & \bf{84.6} & ~\bf{5.2} \\
        \bottomrule
    \end{tabular}
}

}
    \caption{\textbf{Ablation study of the scene-aware mesh recovery module.}
    We validate the upper bound of the proposed method on the RICH dataset. 
    }\label{tab:abl_sahmr}
\end{table}

\PAR{Mesh Recovery Module.}
As shown in Tab.~\ref{tab:abl_parallel}, the parallel network that uses cross-attention outperforms a variant that fuses the pointnet features of the contact points to the METRO network in the early stage.
The CErr indicates the error of contact mesh vertices in the translation-aligned coordinates.
In Tab.~\ref{tab:abl_sahmr}, we validate the upper bound of the mesh recovery module.
We replace the intermediate estimation of root and contact, and find a steady improvement in the pose and shape accuracy that outperform the baseline.

\PAR{Running Time.}
\shortname runs at 170ms with a peak memory cost of 1852 MB for a $224 \times 224$ image and a scene point cloud of $2cm$ resolution on a V100 GPU.
Specifically, the root and contact module takes 92ms (CNN 50ms, SPVCNN 42ms), the mesh recovery module 75ms (CNN 49ms, Transformer 26ms), and 3ms for the intermediate processing.


\section{Conclusion}\label{sec:conclusion}

This work addressed the challenge of estimating the human mesh from an RGB image with the consideration of the scene geometry.
Our key idea is to inject 3D scene cues into a monocular human mesh recovery network to recover the absolute human pose and shape in the scene.
To this end, our approach first predicts the 3D human location and then uses a sparse 3D CNN to estimate dense human-scene contacts.
We developed a transformer to extract features from contact scene points and fed them into the pose estimation network using the cross-attention scheme.
Experiments demonstrated our approach achieves state-of-the-art performance on the RICH and PROX datasets.

\PAR{Acknowledgement.}
This work was supported by NSFC (Grant 62172364) and the Information Technology Center and State Key Lab of CAD\&CG, Zhejiang University.

{\small
\bibliographystyle{unsrt}
\bibliography{fullbib}
}

\end{document}